\documentclass[a4paper, 10pt, conference]{ieeeconf}      
\usepackage{FG2021}
\usepackage{caption}
\usepackage{graphicx}
\FGfinalcopy 

\usepackage{listings}
\usepackage{color}
\usepackage[utf8]{inputenc}

\definecolor{dkgreen}{rgb}{0,0.6,0}
\definecolor{gray}{rgb}{0.5,0.5,0.5}
\definecolor{mauve}{rgb}{0.58,0,0.82}

\IEEEoverridecommandlockouts                              
\def\FGPaperID{0413} 

\title{\LARGE \bf
Solving the Families In the Wild Kinship Verification Challenge by Program Synthesis}

\author{\parbox{16cm}{\centering
    {\large Junyi Huang$^1$, Maxwell Benjamin Strome$^2$,  Ian Jenkins$^3$,  Parker Williams$^2$, Bo Feng$^4$, Yaning Wang$^4$, Roman Wang$^2$, Vaibhav Bagri$^2$, Newman Cheng$^2$,\\ Iddo Drori$^{2,5}$}\\
    {\normalsize
    $^1$ Department of Mathematics, Columbia University\\
    $^2$ Department of Computer Science, Columbia University\\
    $^3$ Department of Applied Physics and Math, Columbia University\\
    $^4$ Department of Electrical Engineering, Columbia University\\
    $^5$ Department of Electrical Engineering and Computer Science, MIT}}
}

\begin{document}

\IEEEoverridecommandlockouts\pubid{\makebox[\columnwidth]{978-1-6654-3176-7/21/\$31.00~\copyright{}2021 IEEE \hfill} \hspace{\columnsep}\makebox[\columnwidth]{ }}

\ifFGfinal
\thispagestyle{empty}
\pagestyle{empty}
\else
\author{Anonymous FG2021 submission\\ Paper ID \FGPaperID \\}
\pagestyle{plain}
\fi
\maketitle


\begin{abstract}
Kinship verification is the task of determining whether a parent-child, sibling, or grandparent-grandchild relationship exists between two people and is important in social media applications, forensic investigations, finding missing children, and reuniting families. We demonstrate high quality kinship verification by participating in the 2021 Recognizing Families in the Wild challenge which provides the largest publicly available dataset in the field. Our approach is among the top 3 winning entries in the competition. We ensemble models written by both human experts and a foundation model, OpenAI Codex, trained on text and code. We use Codex to generate model variants, and also demonstrate its ability to generate entire running programs for kinship verification tasks of specific relationships.
\end{abstract}

\section{INTRODUCTION}
The ability to recognize kinship between faces based only on images presents an important contribution to applications such as social media, forensics, reuniting families, and genealogy. However, these fields each possess unique datasets that are highly varied in terms of image quality, lighting conditions, pose, facial expression, and viewing angle that makes creating an image processing algorithm that works in general quite challenging. To address these issues, an annual automatic kinship recognition challenge \textit{Recognizing Families In the Wild} (RFIW) releases a sizeable multi-task dataset to aid the development of modern data-driven approaches for solving these important visual kin-based problems \cite{robinson2021survey, robinson2016fiw,robinson2018fiw,robinson2020recognizing}. 

We develop and evaluate models for the task of kinship verfication (T-1) in the RFIW 2021 challenge, which entails the binary classification of two pictures' relationship as kin or non-kin \cite{fang2010towards, li2017kinnet, duan2017}. 

The task of kinship recognition was initially tackled over a decade ago \cite{fang2010towards} by extracting low-level image features for classification. Over the years, methods such as transfer learning \cite{xia2011kinship, xia2012understanding}, metric learning \cite{lu2013neighborhood, Wang2017KinshipVO}, and deep learning approaches \cite{lu2013neighborhood,fang2013kinship} have reached accuracy on par with human performance, and serve as baselines for our work. Deep Siamese neural networks utilize two identical neural networks whose outputs are joined to obtain a final prediction. This approach is commonly used to its scalability across multiple tasks. Recent work utilizes enhanced feature fusion along with feature fusion similarity and cosine similarity of measurements to increase model accuracy \cite{yu2021deep}.

Most recently, self-supervised contrastive learning \cite{zhang2021cnu} has been successful in the tackling kinship verification challenge. Self-supervised contrastive learning is an unsupervised framework which extracts deeper structure and representations of faces for kinship verification. The work builds on ArcFace \cite{deng2019arcface} that won of the 2020 RFIW challenge, by extracting facial features which are fed into a multi-layer perceptron that is trained with a contrastive loss \cite{chen2020simple}. To find kin pairs, the features from the network's hidden layers are used to compute a similarity score which activates at a given threshold.

\begin{figure*}
    \centering
    \includegraphics[width=\textwidth]{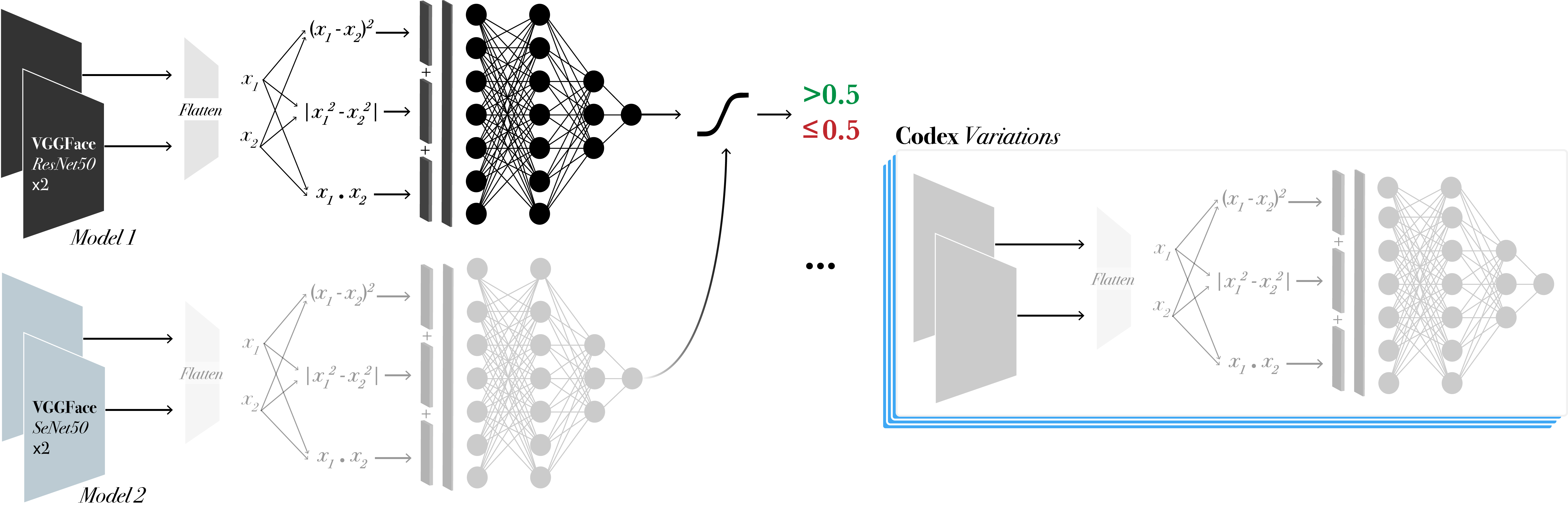}  
    \caption{System architecture: We use multiple deep Siamese networks. A pair of images for verification are fed through a pre-trained convolutional backbone \cite{he2016deep,hu2018squeeze}. The backbones project the images into a latent feature space which are flattened and then combined by feature fusion \cite{yu2020deep}. The result of the feature fusion is fed through a fully connected network in which the final layer is a single binary classification predicting kin or non-kin. We ensemble multiple Siamese networks written by both human experts and OpenAI Codex.} 
    \label{fig:architecture}
\end{figure*}

Our work uses the architecture shown in Figure \ref{fig:architecture} in which a variety of models are written by both human experts and automatically by OpenAI Codex \cite{chen2021evaluating}, a large scale Transformer trained on both text and code. This is the first use of program synthesis to generate a diverse set of neural network models. The models are then ensembled to predict the confidence that a pair of face images are kin. Each model utilizes a Siamese convolutional backbones with pre-trained weights to encode one-dimensional embeddings of each image. We combine the embeddings through feature fusion \cite{he2016deep, hu2018squeeze, yu2020deep}, and feed the fused encoding through a series of fully connected layers in order to make a prediction. The network predictions of many models are ensembled before applying a threshold to obtain a binary classification. We present an ablation study to quantify the impact of our methods starting from a baseline model, measuring the accuracy changes due to each incremental model improvement, while keeping all other factors equal.

\begin{figure}
    \centering
    \includegraphics[width=\columnwidth]{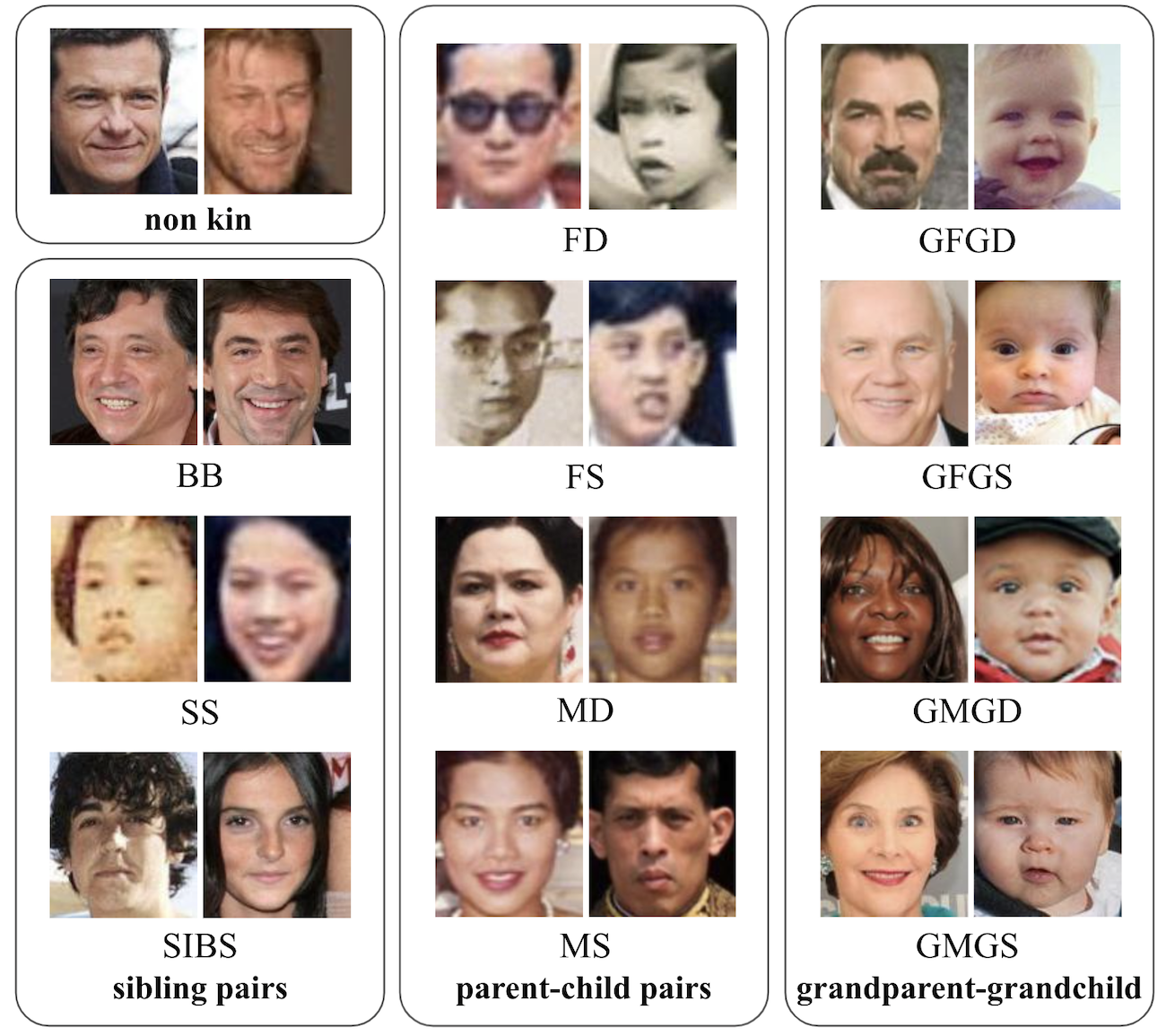}  
    \caption{Example face pairs of each of the 11 relationship type in the FIW dataset: (i) Three sibling pairs: Brother-Brother (BB), Sister-Sister (SS), Brother-Sister (SIBS) (bottom left column); (ii) Four parent-child pairs: Father-Daughter (FD), Father-Son (FS), Mother-Daughter (MD), Mother-Son (MS) (center column); (iii) Four granparent-grandchild pairs: Grandfather-Granddaughter (GFGD), Grandfather-Grandson (GFGS), Grandmother-Granddaughter (GMGD), Grandmother-Grandson (GMGS) (right column); and Non-kin (top left column).}
    \label{fig:relationships}
\end{figure}

\section{METHODS}

\begin{table}[t!]
\small
\centering
\begin{tabular}{|l|c|} 
\hline
\textbf{Model} & \textbf{Accuracy}\\  
\hline
Ensembling multiple instances  &\\ 
of multiple models & 0.741\\
\hline
Ensembling multiple instances &\\ 
of one model & 0.726\\
\hline
Adding $|x_{1}^{2}-x_{2}^{2}|$ to the concatenated features& \\
Test time data augmentation & 0.730\\
\hline
Adding batch normalization layers & \\
and improving sampling & 0.685 \\
\hline
Adding dropout layers & 0.633\\
\hline
Adding training pairs & 0.579\\ 
\hline
Adding a dropout layer & 0.525\\
\hline
Baseline model & 0.510\\
\hline
\end{tabular}
\caption{\label{tab:ablation} Human ablation study: A ranking of different methods that we use to improve our models. Multiple ablation experiments are performed on a smaller dataset consisting of 5,045 training images and 4,437 test images for faster turnaround time. Each improvement in ranking represents an enhancement on top of the prior model. The best performing method consisted of an ensemble of multiple models with different architectures. }
\end{table}

\begin{table*}[t!]
    \scriptsize
    \centering
    \begin{tabular}{|lc|c|ccc|cccc|cccc|} 
    \hline
    \textbf{User} & \textbf{Year} & \textbf{Avg.} & \textbf{BB} & \textbf{SIBS} & \textbf{SS} & \textbf{FD} & \textbf{FS} & \textbf{MD} & \textbf{MS} & \textbf{GFGD} & \textbf{GFGS} & \textbf{GMGD} & \textbf{GMGS}\\
    \hline
zxm123 & 2021 &	0.80 (1)&  0.82 (1)&	0.80 (1)&	0.84 (1)&	0.75 (4)&	0.82 (1)&	0.80 (1)&	0.77 (2)&	0.76 (3)&	0.71 (4)&	0.75 (2)&	0.59 (10)\\
vuvko &	2020 & 0.78 (2)&	0.80 (2)&	0.77 (2)&	0.80 (2)&	0.75 (7)&	0.81 (5)&	0.78 (2) &	0.74 (7) & 0.78 (2)&	0.69 (6)&	0.76 (1)&	0.60 (9)\\
\textbf{nc2893} & 2021 & 0.77 (3)&	0.79 (3)&	0.75 (4)&	0.79 (3)&	0.76 (2)&	0.78 (12)&	0.75 (7)&	0.74 (9)&	0.70 (15)&	0.67 (9)&	0.70 (8)&	0.59 (10)\\
\textbf{jh3450} & 2021 & 0.77 (3)&	0.79 (3)&	0.75 (4)&	0.79 (3)&	0.76 (2)&	0.78 (12)&	0.75 (7)&   0.74 (9)&	0.70 (15)&	0.67 (9)&	0.70 (8)&	0.59 (10)\\
\textbf{paw2140} & 2021 & 0.77 (4)&	0.78 (4)&	0.75 (5)&	0.79 (4)&	0.75 (6)&	0.78 (12)&	0.76 (4)&	0.74 (8)&	0.68 (18)&	0.69 (7)&	0.72 (5)&	0.59 (10)\\
DeepBlueAI & 2020 & 0.76 (5)&	0.77 (5)&	0.75 (6)&	0.77 (5)&	0.74 (8)&	0.81 (6)&	0.75 (6)& 0.74 (10) & 0.72 (7)&	0.73 (3)&	0.67 (11)&	0.68 (1)\\
ustc-nelslip & 2021 & 0.76 (6)&	0.75 (6)&	0.72 (9)&	0.74 (8)&	0.76 (3)&	0.82 (2)&	0.75 (8) & 0.75 (4) & 0.79 (1)&	0.69 (7)&	0.76 (1)&	0.67 (2)\\
\hline
\end{tabular}
    \caption{RFIW kinship verification accuracy scores of top 7 entries in 2020-2021. The Table shows accuracy for each of the 11 types of relationships (three sibling pairs BB, SIBS, SS; four parent-child FD, FS, MD, MS; and four grandparent-grandchild relationships GFGD, GFGS, GMGD, GMGS) and the average accuracy. Our top 3 entries in the competition are shown in bold.}
    \label{tab:results}
\end{table*}

\begin{table*}[t!]
    \centering
    \begin{tabular}{|l|c|ccc|cccc|}
    \hline
    \textbf{Method} & \textbf{Average} & \textbf{BB} & \textbf{SIBS} & \textbf{SS} & \textbf{FD} & \textbf{FS} & \textbf{MD} & \textbf{MS}\\
    \hline
Ensemble & 0.77 & 0.80 & 0.77 & 0.80 & 0.74 & 0.76 & 0.77 & 0.75\\
\hline
Codex model variant 1 & 0.75 & 0.78 & 0.75 & 0.78 & 0.71 & 0.73 & 0.75 & 0.72\\
Codex model variant 2 & 0.76 & 0.80 & 0.76 & 0.79 &	0.72 & 0.76 & 0.76 & 0.72\\ 
\hline
Human model variant 1 & 0.75 & 0.78 & 0.74 & 0.78 & 0.73 & 0.75 & 0.75 & 0.73\\ 
Human model variant 2 & 0.76 & 0.79 & 0.76 & 0.79 & 0.75 & 0.77 & 0.75 & 0.75\\
\hline
\end{tabular}
    \caption{RFIW kinship verification accuracy scores of our ensemble, Codex model variants, and human model variants. Human model variant 1 is based on ResNet50 \cite{he2016deep} and human model variant 2 is based on SENet50 \cite{hu2018squeeze}. Codex model variants consists of multiple fully connected layers written automatically given the human model variants. Finally, we ensemble all four model variations, for our best results.}
    \label{tab:codex-human-results}
\end{table*}

\subsection{Dataset}
Families In the Wild (FIW) \cite{robinson2016fiw} is the largest database for kinship recognition to date. The FIW dataset is split into disjoint training, validation, and test sets. Table \ref{tab:splits} provides our exact splits, with the number of unique faces and families in our dataset.

The test set consists of roughly an equal number of positive and negative examples. For each image, the dataset contains: (i) a binary kinship label, \textit{kin} or \textit{not kin}; (ii) the type of relationship if one exists; (iii) a unique ID for each person in a family, and the families are disjoint.

There are 11 types of relationships in the RFIW challenge dataset split into three overarching groups as shown in Figure \ref{fig:relationships}. The relationships consist of: (i) three sibling pairs: sister-sister (SS), brother-brother (BB), and brother-sister (SIBS); (ii) four parent-child pairs: father-daughter (FD), father-son (FS), mother-daughter (MD), mother-son (MS); and (iii) four grandparent-grandchildren pairs: grandfather-granddaughter (GFGD), grandfather-grandson (GFGS), grandmother-granddaughter (GMGD), grandmother-grandson (GMGS).

\subsection{Data augmentation} 
We apply image transformations to regularize our models and improve their generalization ability. We perform experiments to identify transformations that improve validation and test accuracy. Our best performing model as shown in Table \ref{tab:results} includes applying data augmentation to the input pair. We did not observe a significant change in the generalization error of our models. Further experimentation shows that transformations such as large angle rotations and vertical flips of the images during training time may degrade model performance.  Augmentations such as random small angle rotations, minor crops, horizontal flips, and color channel transformations such as brightness shifts regularize our models, particularly in scenarios where we allow for fine tuning of the backbones. 

In addition to leveraging data augmentation during training, we also introduce test time augmentation \cite{sun2020test}. We evaluate our trained model on the raw test image pair, and also on a variety of augmented versions of this pair. We generate two additional copies of the input pair and perform a horizontal flip on one copy and a color transformation of the other. We then predict the kinship between the two augmented pairs and original pair, and average their confidences. We leave the order of the pair consistent, since our model's feature fusion \cite{yu2020deep} is invariant to the ordering of the images. Other models that are not invariant to the image order may benefit from swapping the images as an additional transformation.

\begin{table}[b!]
\small
\centering
\begin{tabular}{|l|c|c|} 
\hline
\textbf{Split} & \textbf{\# single unique faces} & \textbf{\# of families}\\
\hline
Training & $21,920$ & $571$\\

Validation & $5,045$ & $192$\\

Test & $5,226$ & $190$\\
\hline
Total & $32,191$ & $953$\\
\hline
\end{tabular}
\caption{\label{tab:splits} Dataset splits: Number of unique faces and number of families used for training, validation, and testing.}
\end{table}

\subsection{Architecture}

We utilize a deep Siamese network for kinship verification \cite{yu2020deep}. The deep Siamese network contains two separate branches, where each branch is given one image from the pair selected for verification. Each branch begins with a deep convolutional neural network which projects the image into a latent feature space. The resulting feature vectors from the separate branches are then combined by feature fusion \cite{yu2020deep} and fed into a fully connected network to capture non-linear interactions among the relationship between the two feature vectors. The final layer of the fully connected network is a binary classification which predicts whether a given pair of images is kin or non-kin. 

\subsection{Feature Fusion}
We utilize feature fusion to detect and identify a descriptive set of facial features to improve the accuracy of kinship verification. Each backbone of our architecture produces a 1x$N$ embedding for an input image. Each 1D image embedding ($x_i$) in the pair is then fused with its counterpart by (i) taking the Hadamard product of the feature vectors ($x_1 \cdot x_2$); (ii) the squared difference of the feature vectors $(x_1 - x_2)^2$, and; (iii) the absolute value of the difference of squares ($|x_1^2 - x_2^2|$). 

\subsection{Ensemble}
To increase the generalization ability and robustness of our model, we ensemble the results of a diverse set of network architectures to obtain a final prediction. These architectures include models with different feature extraction backbones to leverage the features learned by disparate network structures. We use ResNet50 \cite{he2016deep}, SENet50 \cite{hu2018squeeze}, FaceNet \cite{schroff2015facenet}, and VGGFace \cite{parkhi2015deep} as backbones.

During training, we also ensemble across different splits of the training data. We first split the data into $k$ folds, selecting one fold as the validation data for networks trained on all other folds. We repeat the process for fold 1 to fold $k$, which results in $k$ ensemble member networks based on one individual backbone model. With 4 backbone models, as mentioned above for instance, $4k$ ensemble member networks are generated for prediction.

\subsection{Sampling}
There are key features of the FIW dataset that make sampling methodologies non-trivial: several people have more pictures than others, and several families have more people than others. Any sampling method must compromise between evenly sampling across pairs of people and families alike. Based on our ablation studies, we found that models perform well when we increase diversity. Therefore, we prioritize sampling evenly across: (i) families; (ii) people; and (iii) pictures, in that ranking order.

\subsection{Test Set Split}
The test set includes roughly the same numbers of positive pairs and negative pairs. We utilize this information by setting an adaptive threshold on the original model outputs which makes our final prediction roughly equally split into positive and negative labels. Knowing the structure of the test set distributions allows us to factor in prior probabilities to our predictions, which slightly improves model performance.

\subsection{Program Synthesis}

\subsubsection{Program Variants}
We leverage program synthesis to improve performance on this challenge by synthesizing architectural components and hyperparameters. To do so we provide prompts including part of our model code to OpenAI's Codex \cite{chen2021evaluating} and incorporate the architectural and hyperparameter changes generated as model variants. These architectural changes written automatically by OpenAI Codex \cite{chen2021evaluating} suggest different combinations for stacking and mixing feature maps in Siamese networks. The same applies to prompting Codex to ensemble multiple models together -- through a series of well-defined sentences provided as prompts, Codex is able to write code for ensembling multiple models together and improves overall performance. Providing guidance through human code snippets allows Codex to solve these tasks by automatically writing variants of existing code. We apply program synthesis to rapidly generate a diverse set of models, and include these variations in our ensemble.

\subsubsection{Entire Programs}
Prompting Codex with:\\
\texttt{"Write a program to compute whether two faces are related."},\\
results in code that also transforms the images, rotating them based on the eyes and eyebrows.\\
Prompting Codex with specific tasks such:\\
\texttt{"Predict whether a father/son relation exists between two faces."}\\ 
or specifying other relationships such as "mother/daughter" or "grandfather/grandson". Codex generates code that utilizes SiameseNet \cite{bromley1993signature}, and ContrastiveLoss \cite{zhang2021cnu}.\\
Prompting Codex to use specific platforms or software packages: \texttt{"Write an ensemble network using Tensorflow that determines whether a father/son relationship exists between two face images. Use VGG16 and ResNet in the model."}\\
results in code which builds a model using VGG16 \cite{simonyan2014very} and ResNet \cite{he2016deep} as the backbone without including the classification layers at the top, allowing for feature extraction and fine-tuning. See our project page for specific code examples.

\section{RESULTS}

Table \ref{tab:results} contains the result of our best performing model trained on the 2021 RFIW kinship verification dataset shown by the bold entry in the user column. We compare our results to the top submissions from 2020-2021. Our model performs in the top three overall. In Table \ref{tab:codex-human-results}, the prediction accuracy of four ensemble networks and a super ensemble model are compared. Human variant 1 and Codex variant 1 are constructed with ResNet50 \cite{he2016deep} as the backbone model, while human variant 2 and Codex variant 2 use SENet50 \cite{hu2018squeeze} as the backbone model. 

The Codex variants contain architectural modifications of human variants with more fully connected layers after the Siamese networks. Multiple instances of each model are trained by k-fold cross-validation. All models are ensembled to form a model which performs best overall. 

We perform an ablation study as shown in Table \ref{tab:ablation}, on the number of dropout layers, batch normalization layers, addition of difference of squares in feature fusion, sampling techniques, test time augmentation, and model ensembling. The results show that all the proposed architectural components improve upon the baseline. The strongest results is achieved by ensembling multiple instances of multiple models, which is our full model.

\section{CONCLUSIONS}
This work achieves a top 3 position in the 2021 kinship verification challenge over all years. We use a base Siamese network architecture for predictions. Our top performing model is an ensemble of diverse models which utilize training and test time data augmentation. Our work is the first to use models written both human experts and written automatically by OpenAI Codex \cite{chen2021evaluating}. This work provides a strong result on the important kinship verification task, and perhaps most importantly demonstrates that a hybrid human-machine approach may advance the field. We hope this work opens the door to the use of program synthesis in other common task framework computer vision challenges.


{\small
\bibliographystyle{ieee}
\bibliography{bibliography}
}


\end{document}